# VisioBlend: Sketch and Stroke-Guided Denoising Diffusion Probabilistic Model for Realistic Image Generation

Mr. Harshkumar Devmurari
*Dept. of CSE(DS)*
*Vidyavardhini's College of Engineering and Technology*
Mumbai, India
harshdevmurari007@gmail.com

Mr. Gautham Kuckian
*Dept. of CSE(DS)*
*Vidyavardhini's College of Engineering and Technology*
Mumbai, India
gauthamkuckian07@gmail.com

Mr. Prajjwal Vishwakarma
*Dept. of CSE(DS)*
*Vidyavardhini's College of Engineering and Technology*
Mumbai, India
prajjwalvishwakarma360@gmail.com

Ms.. Krunali Vartak
*Assistant Professor, Dept. of CSE(DS)*
*Vidyavardhini's College of Engineering and Technology*
Mumbai, India
krunali.vartak@vcet.edu.in

*Abstract-* **Generating images from hand-drawings is a crucial and fundamental task in content creation. The translation is challenging due to the infinite possibilities and the diverse expectations of users. However, traditional methods are often limited by the availability of training data. Therefore, VisioBlend, a unified framework supporting three-dimensional control over image synthesis from sketches and strokes based on diffusion models, is proposed. It enables users to decide the level of faithfulness to the input strokes and sketches. VisioBlend achieves state-of-the-art performance in terms of realism and flexibility, enabling various applications in image synthesis from sketches and strokes. It solves the problem of data availability by synthesizing new data points from hand-drawn sketches and strokes, enriching the dataset and enabling more robust and diverse image synthesis. This work showcases the power of diffusion models in image creation, offering a user-friendly and versatile approach for turning artistic visions into reality.**

## I. INTRODUCTION

Sketches and strokes serve as vital communication mediums, representing abstract depictions of objects and scenes that reflect diverse human creations and ideas. Translating these abstract illustrations into tangible images can bridge human creativity with reality, unlocking a realm of potential applications and aiding content creation processes. However, the task of image generation from sketches and strokes is inherently complex due to its ill-defined nature[1]. Different users may expect varying outputs under different circumstances, with preferences for the degree of faithfulness differing widely.

Traditionally, this problem has been approached as image-to-image translation using generative adversarial networks (GANs). However, existing methods are often task-dependent, requiring separate models for different translation tasks and lacking flexibility and controllability in terms of faithfulness. Recent advancements in diffusion models have shown promise in high-quality image synthesis with stable training procedures, offering a more flexible and controllable approach[2][3].

In this, we propose a unified framework called VisioBlend (DIffusion-based Sketch and Stroke synthesis), which allows for three-dimensional control over the synthesis process, specifically targeting contour, color, and realism. Unlike previous works that handle either black-white sketches or stroke paintings, VisioBlend simultaneously addresses both factors, offering disentangled control for shape and color consistency. This is achieved through classifier-free guidance, enabling two-dimensional control and customization of the generative process based on user demands.

Furthermore, to address the challenge of input sketches and strokes being inconsistent with the distribution of real images, we introduce a third control factor: the realism scale. This scale allows users to trade off between consistency and realism, enhancing the flexibility of the synthesis process[4]. By utilizing iterative latent variable refinement and a low-pass filter, VisioBlend adjusts the coarse-to-fine features of the referred drawings, enhancing the realism of the generated images.

VisioBlend facilitates various applications, including multi-domain translation, multi-conditioned local editing, and region-sensitive stroke-to-image synthesis. We evaluate the framework quantitatively and qualitatively on three-dimensional controllability, demonstrating its effectiveness in generating realistic images while allowing users to customize the degree of faithfulness to input sketches and strokes.

## II. LITERATURE REVIEW

### A. Overview Of Existing Research

The research [5], proposes a novel face photo retrieval system using sketch drawings, focusing on transforming a photo image into a sketch to enable effective matching. It separates shape and texture information in face photos and conducts transformation on them separately, achieving high efficacy in face sketch recognition.

In [6], it presents a sketch-based image retrieval algorithm that focuses on measuring the similarity between a sketch and an image. It proposes a method called salient contour reinforcement, which divides the image contour into global and salient contour maps. These maps are used to extract a new descriptor called angular radial orientation partitioning (AROP) feature, which utilizes edge pixels' orientation information to identify spatial relationships.

The research [7], proposes a method for improving face sketch to photo matching by addressing shape exaggeration and illumination variation issues. It introduces new fiducial points for geometric face alignment to reduce shape exaggeration and describes images using a novel feature descriptor called Difference of Gaussian-Oriented Gradient Histogram (DoGOGH) to minimize illumination effects.

In [8], it proposes a contextual generative adversarial network (GAN) for image generation guided by hand-drawn sketches. It addresses the problem of aligning input sketches with output images by using the sketch as a weak constraint for image completion. The proposed approach outperforms state-of-the-art conditional GANs on challenging inputs and generalizes well on common categories.

Research [9] presents a convolutional neural network (CNN) semantic re-ranking system for enhancing sketch-based image retrieval (SBIR) performance. The system leverages CNNs to capture category information from sketches and natural images, enabling effective similarity measurement. By training dual CNN models, the system achieves accurate classification of query sketches and natural images. The proposed re-ranking method uses category similarity measurement to re-rank initial retrieval results, resulting in significantly higher precision compared to baseline systems and other re-ranking algorithms on various SBIR datasets.

In [10], it presents an interactive approach for line art colorization using conditional Diffusion Probabilistic Models, improving creative productivity. It outperforms existing methods in generating diverse and high-quality colorized images, evaluated using FID, LPIPS, and SSIM metrics.

In [11], proposes a novel controllable sketch-to-image translation framework for synthesizing and editing face images from hand-drawn sketches. The framework includes a dilation-based sketch refinement method and leverages scale-aware style transfer for multi-level refinement, allowing users to control the fidelity of the output. Advanced user controllability features, such as facial attribute editing and spatially non-uniform refinement, are also explored, demonstrating the effectiveness of the proposed method through extensive experiments and comparisons with state-of-the-art methods.

The research in [12] a novel deep learning architecture proposed for zero-shot sketch-based image retrieval. The paper reviews traditional and deep learning-based methods, highlighting challenges like the semantic gap and data scarcity. TCN addresses these by aligning sketch and image features and learning transferable representations. The model's ability to adapt to new categories enhances its practicality. TCN's introduction signifies a significant advancement in zero-shot retrieval, promising more accurate and robust retrieval systems for sketch-based image search.

The [13], introduces a two-step method and interactive system for architecture sketch-to-image translation, overcoming challenges of large domain gaps and information asymmetry. The method includes a supervised GAN for generating gray sketches from binary sketches and an unsupervised GAN for translating gray sketches into various color images, outperforming existing methods in both image quality and diversity.

### B. Characteristics Of realistic image from sketch and stroke

1) Unified framework: Unlike traditional methods that require separate models for different tasks (e.g., sketch-to-image and stroke-to-image translation), VisioBlend provides a unified framework that can handle both sketches and strokes simultaneously [14]. This unified approach not only simplifies the model architecture but also improves efficiency and flexibility in generating images from different types of inputs.

2) Diffusion-based framework: The framework is built upon diffusion models, which are a class of generative models known for their ability to generate high-quality images. Diffusion models operate by iteratively adding noise to an image and then denoising it, effectively learning a probability distribution over images. This approach leads to stable training and the generation of realistic images.

3) Flexible editability: VisioBlend enables flexible editing of images based on real images by simply drawing contours and colors. This feature allows users to make quick and easy edits to the generated images, such as changing the color of an object or adding new details, without the need for complex editing tools. This flexibility makes VisioBlend ideal for content creation and customization.

4) Performance: VisioBlend has been shown to achieve state-of-the-art performance in terms of image synthesis quality and stability [15]. This is demonstrated through quantitative metrics such as Frechet Inception Distance (FID) and LPIPS, which measure the realism and perceptual quality of the generated images. Qualitative user studies further validate the effectiveness of VisioBlend in producing high-quality images that are faithful to the input sketches and strokes.

5) Applications: VisioBlend demonstrates its versatility through various applications. For example, it can be used for translation, allowing users to translate sketches into different styles or domains [16]. It can also be used for multi-conditioned local editing, enabling users to edit existing images by drawing contours and colors. Additionally, it supports region-sensitive stroke-to-image generation, providing variations on blank regions in the input sketches and strokes.

## C. Analysis And Limitations Of the Existing System

The limitations of the existing papers system are:

1) Limited scalability: Existing systems often face challenges in scaling up to handle a large number of classes or complex datasets. This limitation can affect their practical utility in scenarios where a wide range of image categories need to be classified or generated. As the number of classes increases, the computational and memory requirements of these systems may become prohibitive, leading to performance degradation or increased inference times[17].

2) Lack of generalization: Some systems may struggle to generalize well to unseen or out-of-distribution classes. This limitation can arise due to the inherent biases in the training data or the complexity of the model architecture. When faced with novel classes or data distributions, these systems may exhibit reduced performance, limiting their applicability in real-world scenarios where such challenges are common.

3) Interpretability challenges: The inner workings of complex generative models like GANs and VAEs can be difficult to interpret. This lack of interpretability can hinder users' understanding of how the model generates images and make it challenging to control or manipulate the generation process effectively. Without a clear understanding of the model's decision-making process, users may struggle to diagnose and correct errors or biases in the generated outputs.

4) Mode Collapse: GAN-based approaches are susceptible to mode collapse, where the generator fails to produce diverse outputs, instead converging to a few modes in the data distribution[18].

5) Limited controllability: While some systems offer control over certain aspects of image generation, such as style or color, achieving fine-grained control over specific attributes like pose or expression remains challenging. This limitation can restrict the practical utility of these systems in applications where precise control over image attributes is desired, such as in fashion design or digital art.

6) Robustness to Input Variability: Some methods may not perform well when faced with highly variable or noisy inputs, limiting their effectiveness in scenarios where input quality varies.

## III. METHODOLOGY

3.1 Model Architecture

Our sketch and stroke-guided diffusion model utilizes a modified U-Net architecture for posterior prediction within the diffusion process. This section details the core components of our model:

### A. Base U-Net Architecture

The foundation of our model is a U-Net architecture as described earlier. The U-Net is a convolutional neural network (CNN) well-suited for tasks involving image segmentation and generation. It employs a symmetrical structure comprising two main parts:

1. Encoder: This part consists of a sequence of residual layers and downsampling convolutions. The residual layers help preserve information flow through the network, while the downsampling convolutions progressively reduce the spatial resolution of the feature maps. This process aims to extract high-level features from the input data.
2. Decoder: A sequence of residual layers and upsampling convolutions is employed. The upsampling convolutions increase the spatial resolution of the feature maps, allowing for reconstruction of the image. Skip connections directly link corresponding layers in the encoder and decoder with the same spatial size, thus leading to improved image generation quality.

### B. Latent Diffusion Model with U-Net Encoder and Decoder

Our model incorporates a latent diffusion model (LDM) as the core diffusion process. Latent diffusion models (LDMs) are a specific type of diffusion model that operates in the latent space of the data. The latent space is a lower-dimensional representation of the original data that captures its essential characteristics.

Advantages of Latent Diffusion Models

1. Higher Fidelity: By working in the latent space, LDMs can capture more complex and intricate image features, leading to higher-fidelity image generation.
2. Better Conditioning: LDMs can effectively incorporate additional conditioning information, such as sketch and stroke data in our case, within the latent space. This allows for more precise control over the generated image content.
3. Improved Efficiency: LDMs can often achieve better results with fewer diffusion steps compared to standard diffusion models, making them computationally more efficient.

When dealing with image generation, LDE proves to be more robust and lightweight.

### C. Modified U-Net for Sketch and Stroke Guidance

To incorporate sketch and stroke information, we modified the standard U-Net architecture by extending the input channel from 3 (for RGB images) to 7.

Initially the input to LDE U-Net at any given timestamp was just the noisy image predicted at previous timestamp with some additional information. The image that is provided as input is generally of dimension (h,w,3). We modified the architecture for posterior probability prediction not only through noisy images but also incorporating the sketch and stroke information. We do this via appending the sketch guidance through 1 greyscale colour channel and stroke information through 3 rgb channels.

Unlike traditional LDE where input is (h,w,3), this channel appending process changes our input layer dimensions to (h,w,7)

7 (Modified Input) = 3 ($X_t$) + 1 ($C_{sketch}$) + 3 ($C_{stroke}$)

where,

Input Image ($X_t$): The current noisy image estimate within the diffusion loop.

Sketch ($C_{sketch}$): A single-channel black-and-white image representing the user-provided sketch.

Stroke ($C_{stroke}$): A 3-channel colored image representing the user-provided strokes.

3.2 Data Preparation

This section details the process of preparing training data for our sketch- and stroke-guided diffusion model. We utilized three publicly available datasets for training:
A. AFHQ-Cat: This dataset consists of a large collection of high-resolution images containing faces of various cat breeds.
B. Flowers: This dataset provides a diverse collection of flower images with varying colors, species, and backgrounds.
C. Landscapes: This dataset offers a wide range of landscape images encompassing different geographical features and natural scenes.

Sketch and Stroke Data Generation:

For Training Datasets:

We employed a pre-trained Photo-sketching model [13] to extract sketch data ($C_{sketch}$) from the images within each dataset. This model, based on Generative Adversarial Networks (GANs), is adept at generating contour drawings from real images.

For the Flowers dataset specifically, we performed an additional foreground extraction step using the GrabCut algorithm in OpenCV [OpenCV reference] before applying the Photo-sketching model. This step aimed to isolate the main flower object from background clutter, which could potentially lead to inaccurate sketch generation.

For Stroke Data:

To represent the strokes guiding the image generation process, we utilized two state-of-the-art image-to-painting frameworks:
A. Stylized Neural Painting [19]: This framework leverages deep learning to translate images into artistic paintings with various styles. We employed Stylized Neural Painting for the AFHQ-Cat dataset due to its effectiveness in capturing artistic brushstrokes.
B. Paint Transformer [20]: This framework utilizes a transformer-based architecture for efficient image-to-painting translation. We employed a Paint Transformer for the Flowers and Landscapes datasets due to its computational efficiency, particularly relevant for processing larger datasets.

Custom Input Image Handling:

For scenarios where users provide custom input images (ccomb) that differ from the training datasets, we adopted a separate approach to extract sketch and stroke data:
A. Foreground Extraction (GrabCut): We utilized the GrabCut algorithm to extract the foreground object (likely the main focus of the image) from the custom input. This step removes background distractions that could hinder sketch and stroke generation [21].
B. Sketch Generation (Canny and findContours): We employed the Canny edge detection algorithm to identify edges within the foreground object. Following this, we utilized the findContours function in OpenCV to extract contour information from these edges. This contour information forms the black-and-white sketch (csketch).
C. Stroke Generation (Contour Coloring): We generated the colored strokes (cstroke) by simply making the contour pixels white within the original input image (ccomb). This approach creates a basic colored outline representing the strokes.

We ensured that our model was trained on a diverse set of images with corresponding sketch and stroke information, allowing it to effectively leverage this guidance to generate realistic images for both pre-defined datasets and user-provided custom inputs.

3.3 Training Procedure

Stage 1: Training with Complete Sketch and Stroke Data:

In the initial stage, the model is trained using the prepared training data where both sketch (csketch) and stroke (cstroke) information is available for each image.

During this stage, the model learns to effectively utilize this complete guidance to generate realistic images that align with the provided sketch and strokes.

Stage 2: Fine-tuning with Partial Information

In the second stage, we employ a fine-tuning strategy to enhance the model's ability to handle scenarios with incomplete guidance.

For a portion of the training data, some elements of the sketch or stroke data are replaced with gray pixels. This simulates situations where users might provide incomplete sketch or stroke information.

By training with this partially obscured data, the model learns to become more robust and generate reasonable results even when lacking complete guidance.

3.4 Evaluation Metrics

To assess the quality of the images generated by our sketch- and stroke-guided diffusion model, we employed two common metrics:

- A. Fréchet Inception Distance (FID): This metric measures the statistical similarity between two datasets of images. In our case, it compares the distribution of the generated images to the distribution of real images from the test set. Lower FID scores indicate better performance, as they suggest the generated images are statistically closer to real images.
- B. Learned Perceptual Image Patch Similarity (LPIPS): This metric evaluates the perceptual similarity between two images. It utilizes a pre-trained deep learning model to assess how similar the images appear to the human visual system. Lower LPIPS scores indicate greater perceptual similarity between the generated images and real images.

By considering both FID and LPIPS, we gain a comprehensive evaluation of the generated images, as FID analyzes how well the model captures the overall data distribution of real images while LPIPS assesses how visually realistic and perceptually similar the generated images are to real images.

These metrics provide quantitative benchmarks for comparing the performance of our model with alternative approaches for sketch- and stroke-guided image generation.

## IV. RESULTS

This section presents the quantitative results obtained by evaluating our sketch- and stroke-guided diffusion model for image generation. We assessed the quality of the generated images using two established metrics: Fréchet Inception Distance (FID) and Learned Perceptual Image Patch Similarity (LPIPS).

*A. Performance on Benchmark Datasets*

We evaluated our model's performance on three publicly available benchmark datasets: AFHQ-Cat (animal faces), Flowers, and Landscapes (LHQ). The model was trained on these datasets with corresponding sketch and stroke information for each image. Table 1 summarizes the FID and LPIPS scores achieved on the test sets of each dataset.

| Dataset | FID | LPIPS |
|---|---|---|
| AFHQ-Cat | 24.02 | 0.172 |
| Flowers | 97.23 | 0.212 |
| LHQ (Landscapes) | 79.71 | 0.183 |

Table 1: FID and LPIPS Scores on Benchmark Datasets

Lower FID scores indicate better performance, as they suggest the generated images for each dataset are statistically closer to the real images within the test set.

Lower LPIPS scores indicate greater perceptual similarity between the generated images and real images, meaning the generated images appear more realistic and natural to the human eye.

Key Observations: Our model achieved the best performance on the AFHQ-Cat dataset, with an FID score of 24.02 and an LPIPS score of 0.172. This suggests that the model was able to effectively capture the statistical distribution and generate images with high perceptual similarity to real cat faces based on the provided sketch and stroke guidance.

The FID scores for the Flowers and LHQ datasets are higher (97.23 and 79.71, respectively) compared to AFHQ-Cat. This might indicate that the model encountered more challenges in generating these more complex and diverse image types (flowers and landscapes) while maintaining statistical similarity to the real data. However, the LPIPS scores for both datasets (0.212 and 0.183) are still relatively low, suggesting that the generated images appear visually realistic despite the slightly higher FID values.

*B. Qualitative Results*

To complement the quantitative evaluation, here we present a set of visual examples showcasing the image generation capabilities of our sketch- and stroke-guided diffusion model. These images were generated for different datasets and with varying sketch and stroke styles.

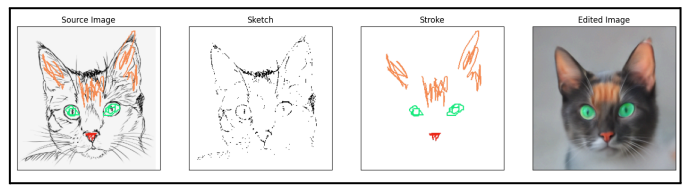

Fig 1: Image Generated with sketch and stroke

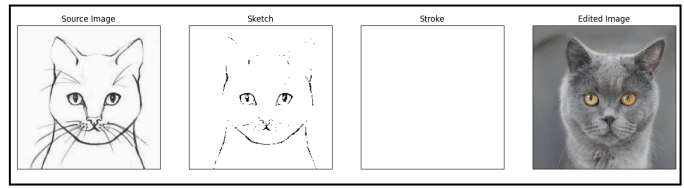

Fig 2: Image Generated with only sketch

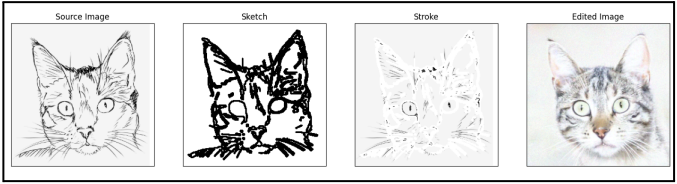

Fig 3: Image Generated with heavy sketch and slight stroke

The above shows the excellent ability of our model to turn sketch into realistic image generation using guided sketch and stroke input. It overall is efficient in generation and performs quite well in scenarios given both sketch and stroke, only sketch no stroke, only stroke no sketch and no sketch and no stroke.

The given below is a friendly UI made for ease of testing and usage.

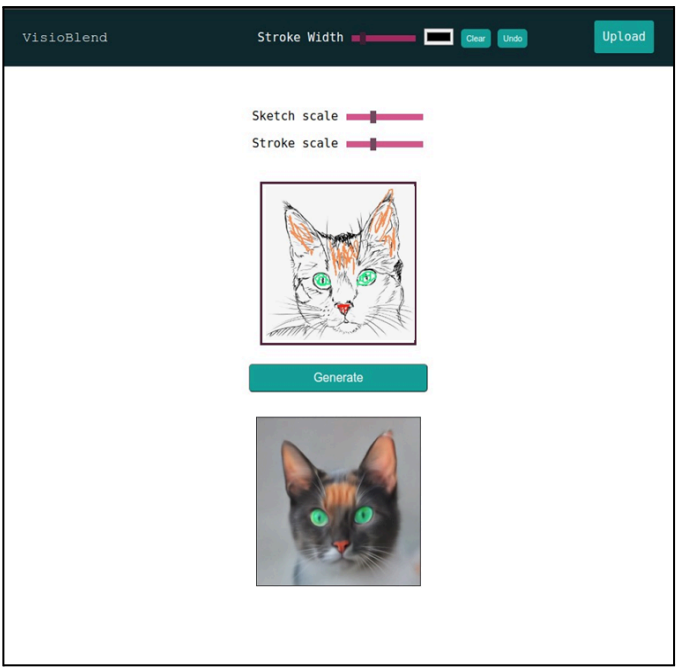

Fig 4: Sketching Canvas Console

These qualitative results visually demonstrate the effectiveness of our model in generating realistic and diverse images across different categories (cats, flowers, landscapes) while incorporating user-provided sketch and stroke information for style and content control.

## V. DISCUSSION

### A. Limitations And Challenges Of the Research

Some potential limitations and challenges that may be associated with research:

1) Complexity: The framework comprises several interconnected components, including stroke extraction, sketch synthesis, and image generation modules. Coordinating these components effectively can be challenging, especially when integrating the framework into existing systems or workflows[19]. The complexity of the framework may require specialized expertise to implement and maintain, potentially limiting its accessibility to users without a strong background in computer vision and machine learning.

2) Computational Resources: Training and running the VisioBlend model can be computationally intensive, particularly when working with high-resolution images or complex datasets. This reliance on computational resources may restrict the scalability of the framework, making it less suitable for deployment in environments with limited computational capabilities or for real-time applications where low latency is crucial.

3) Data Requirements: The effective training of the VisioBlend model typically requires a large volume of annotated data, including pairs of sketches or strokes and their corresponding images[20]. Acquiring and labeling such datasets can be time-consuming, labor-intensive, and expensive, especially for domains or applications with specific requirements or constraints. Limited availability of suitable training data could hinder the model's performance and generalization ability.

4) Robustness to Input Variability: The VisioBlend framework may struggle to maintain performance and consistency when faced with input sketches or strokes that deviate significantly from the training distribution. Handling such input variability while maintaining the desired level of control over image synthesis poses a significant challenge for the framework.

5) Control and Interpretability: While the VisioBlend framework offers some degree of control over the output images' characteristics, such as the level of consistency with the input sketch or stroke, the interpretability of the model's internal representations and the mechanisms governing these controls are not always clear. Improving the interpretability of the model could enhance users' ability to manipulate and fine-tune its behavior, leading to more predictable and desirable outcomes[21].

6) Evaluation Metrics: Evaluation metrics such as FID (Frechet Inception Distance), LPIPS (Learned Perceptual Image Patch Similarity), and user studies are crucial for assessing the quality, fidelity, diversity, and perceptual appeal of generated images in image synthesis tasks. These metrics provide quantitative and qualitative insights into the performance of the model and help researchers understand its strengths and limitations.

7) Controllability and realism trade-off: Balancing the degree of controllability and realism in image synthesis poses a significant challenge. Users may desire precise control over image attributes while expecting realistic outputs, which can be difficult to achieve simultaneously

### B. Suggestions For Future Research

Some potential suggestions for future research in the field of VisioBlend are:

*1) Enhanced Control and Customization:* Future research could focus on developing advanced controls for users to manipulate various aspects of image generation. This includes more granular adjustments for style, color, texture, and other visual elements. Enabling users to set specific constraints or preferences could lead to more personalized and tailored image outputs, enhancing the overall user experience and creative possibilities.

*2) Integration with Existing Workflows:* Research could explore ways to seamlessly integrate the VisioBlend framework into existing creative workflows and software tools. This might involve developing standardized APIs or plugins that allow for easy communication between VisioBlend and other applications. By streamlining the integration process, users can incorporate VisioBlend into their existing workflows without disruptions, increasing adoption and usability.

*3) Collaborative and Social Features:* Future work could focus on enhancing the collaborative and social aspects of the VisioBlend framework. This could include features that allow users to easily share their work, collaborate with others in real-time, and remix each other's creations. By fostering a collaborative environment, VisioBlend can become a hub for creative exchange and community engagement, enriching the overall user experience.

*4) Accessibility and Inclusivity:* Research could aim to make the VisioBlend framework more accessible and inclusive to a wider range of users. This could involve designing user interfaces that are intuitive and easy to navigate, providing comprehensive documentation and tutorials in multiple languages, and considering the diverse needs and preferences of users with disabilities or limited technical expertise. By prioritizing accessibility, VisioBlend can ensure that all users can fully participate in the creative process.

*5) Scalability and Performance:* Future research could focus on improving the scalability and performance of the VisioBlend framework. This includes optimizing algorithms to handle larger datasets and more complex models, leveraging parallel computing techniques to speed up processing times, and exploring new architectures that can enhance the efficiency of image generation processes. By improving scalability and performance, VisioBlend can accommodate the growing demand for high-quality image synthesis in various applications.

## VI. CONCLUSION

The VisioBlend framework represents a significant advancement in the field of image generation from sketches and strokes. It offers a novel approach that enables users to use the degree of consistency between input sketches and generated images, providing a high level of realism. Novel architecture demonstrates the model's ability to generate realistic and diverse images across different categories. This framework has the potential to revolutionize how artists, designers, and creators approach image synthesis, offering a more intuitive and customizable process. It Highlights the potential of sketch and stroke guidance for user-controlled image generation. Despite its strengths, the VisioBlend framework also faces several limitations and challenges. These include the need for more diverse training data to handle a wider range of styles and scenes, as well as improvements in user interface design to enhance usability and accessibility. Despite these challenges, the VisioBlend framework shows great promise for the future of image synthesis and has the potential to inspire new research directions and innovations in the field.